\documentclass{article}
\usepackage{bm, bbm}
\usepackage{amsmath, amsfonts}
\usepackage{mathtools}
\mathtoolsset{showonlyrefs=true}

\newcommand{\argmin}{\mathop{\rm arg~min}\limits}
\newcommand{\rank}{\operatorname{rank}}
\newcommand{\st}{\mathrm{s.t.}}
\newcommand{\defeq}{:=}
\newcommand{\eqdef}{=:}

\usepackage{amsthm}
\newtheorem{theorem}{Theorem}[section]

\usepackage[round,authoryear]{natbib}

\usepackage{authblk}
\makeatletter
\renewcommand\AB@authnote[1]{\rlap{\textsuperscript{\normalfont#1}}}

\makeatother

\title{
	Binary Matrix Completion Using Unobserved Entries
	}
		
\author[1]{Masayoshi Hayashi}
\author[2,3]{Tomoya Sakai}
\author[3,2,1]{Masashi Sugiyama}
\affil[ ]{
	{\{hayashi@ms., sakai@ms., sugi@\}k.u-tokyo.ac.jp}
	\vspace{4mm}
	}

\affil[1]{
  Department of Computer Science, \protect\\
  The University of Tokyo, \protect\\
   7-3-1 Hongo, Bunky\=o-ku, Tokyo, Japan
  \vspace{3mm}
}
\affil[2]{
  Department of Complexity Science and Engineering, \protect\\
  The University of Tokyo, \protect\\
   5-1-5 Kashiwanoha, Kashiwa-shi, Chiba, Japan
  \vspace{3mm} 
}
\affil[3]{
  Center for Advanced Intelligence Project, \protect\\
  RIKEN, \protect\\
   1-4-1 Nihombashi, Ch\=u\=o-ku, Tokyo, Japan
}

\date{}	

\begin{document}
\sloppy
\maketitle

\begin{abstract}
A matrix completion problem, which aims to recover a complete matrix from its partial observations, is one of the important problems in the machine learning field and has been studied actively.
However, there is a discrepancy between the mainstream problem setting, which assumes continuous-valued observations, and some practical applications such as recommendation systems and SNS link predictions where observations take discrete or even binary values.
To cope with this problem,
Davenport et al.\ (2014) proposed a binary matrix completion (BMC) problem, where observations are quantized into binary values.
Hsieh et al.\ (2015) proposed a PU (Positive and Unlabeled) matrix completion problem, which is an extension of the BMC problem.
This problem targets the setting where we cannot observe negative values, such as SNS link predictions.
In the construction of their method for this setting, they introduced a methodology of the classification problem, regarding each matrix entry as a sample.
Their risk, which defines losses over unobserved entries as well, indicates the possibility of the use of unobserved entries.
In this paper, motivated by a semi-supervised classification method recently proposed by Sakai et al.\ (2017), we develop a method for the BMC problem which can use all of positive, negative, and unobserved entries, by combining the risks of Davenport et al.\ (2014) and Hsieh et al.\ (2015).
To the best of our knowledge, this is the first BMC method which exploits all kinds of matrix entries.
We experimentally show that an appropriate mixture of risks improves the performance.
\end{abstract}

{\small
\textbf{Keywords:}
Matrix completion, 
Binary matrix completion, 
Learning from positive and unlabeled data
}

\section{Introduction}
A matrix completion problem, which aims to recover a complete matrix from its partial information, is an important problem in machine learning and has been well studied~\citep{Fazel02, Candes09, Daven16}.
It has been applied to a wide variety of practical problems such as collaborative filtering~\citep{Gold92}, system identification~\citep{Liu09}, sensor localization~\citep{Biswas06}, and rank aggregation~\citep{Gleich11}.
Recently many theoretical analyses on the matrix completion have been conducted~\citep{Recht11, Cai16, Rong16}, and they typically guarantee the accurate recovery of the target matrix under a sufficient number of observed entries.

A mainstream approach to the matrix completion problem assumes continuous-valued observations.
However, there is a discrepancy between those problem settings and real-world applications.
There are some real-world applications, whose observations take discrete or even binary values.
For instance, in a famous collaborative filtering problem of the Netflix Prize~\citep{netflix, netflix_workshop}, the input ratings take integer values from $1$ to $5$.
Also, in the YouTube rating system, there are only two values (``good'' and ``bad'') for input.
These quantized observations can be considered to be generated based on some underlying real-valued matrix, but it is difficult to estimate this matrix without considering how the quantization occurs.

To cope with such problems, \citet{Daven14} proposed a new problem setting called the binary matrix completion (BMC) problem, where observations take binary values.
They also demonstrated the superior performance of their method in the experiment with a movie rating dataset.
Although their setting can handle problems with binary quantized observations, there are more difficult situations in practice.
For example, in some social networking services, we only observe ``like'' as a response to web articles
and articles without responses are not directly considered as ``unlike'' but can be either ``like'' or ``unlike''.

This type of problem is called \emph{learning from positive and unlabeled data} (PU learning) and has been widely studied in the classification field~\citep{Elkan08, Niu16, Kiryo17}.
However, there is a gap between the classification and matrix completion problems, for instance, unlabeled samples do not exist in the context of matrix completion.
To address the PU matrix completion problem, where there are no negative observations, \citet{Hsieh15} regarded unobserved entries as unlabeled data.
Their PU matrix completion method is based on well-studied PU learning techniques in classification tasks~\citep{Hsieh15}.

An advantage of the PU matrix completion method is that it can take unobserved entries into account, unlike other existing matrix completion methods which use only observed entries.
This suggests that we can utilize unobserved entries for estimating matrices, also in the BMC problem, where both positive and negative entries are observable.
One way to achieve that would be to extend the PU matrix completion method so that it can also handle negative observations.

In this paper, we propose a novel BMC method by incorporating unobserved entries in addition to positive and negative observations.
More specifically, we combine the PU matrix completion method~\citep{Hsieh15} and its counterpart, the NU (negative and unlabeled) matrix completion method, with a BMC method.
Our approach is motivated by a semi-supervised learning method based on the PU learning for classification tasks~\citep{Sakai17}.
Their idea is to combine a supervised learning method with a PU learning method so that it can utilize unlabeled data for learning classifiers.

In the work of \citet{Sakai17}, the authors considered combinations of supervised learning, PU learning, and NU learning, and concluded that combinations of supervised learning with either PU or NU learning is a promising approach from the viewpoint of both theory and empirical results.
To extend the PU matrix completion method to handle negative entries, we also need to investigate promising combinations for the BMC problem setting.
Since the matrix completion problem is substantially different from the classification problem considered in \citet{Sakai17}, we discuss an appropriate approach in the context of matrix completion.

The rest of this paper is structured as follows.
In Section~\ref{ch:pre}, we first introduce notations and then formulate the binary matrix completion problem.
In Section~\ref{ch:mc} and Section~\ref{ch:class}, we review existing matrix completion and PU classification methods.
In Section~\ref{ch:proposed}, we propose a binary matrix completion method based on a BMC method and a PU matrix completion method.
In Section~\ref{ch:exp}, we experimentally analyze our approach and demonstrate the effectiveness of the proposed method on benchmark datasets.
Finally, in Section~\ref{ch:conclusion}, we conclude the paper and discuss our future work.

\section{Preliminaries}\label{ch:pre}
In this section, we introduce notations we use in this paper and then give a problem setting of the binary matrix completion problem.

\subsection{Notations}
For any positive integer $n$, denote $\{1,2,3,\ldots,n\}$ by $[n]$.
For any pair of real numbers $a$ and $b$, define $a \vee b \defeq \max(a,b)$ and $a \wedge b \defeq \min(a,b)$.
We use $\mathbb{S}^d$ and $\mathbb{S}^{d_1\times d_2}$ to denote a space of all $d$-dimensional vectors and all $d_1\times d_2$-matrices, which consist of elements of a set $\mathbb{S}$, respectively.
For example, ${\{0,1\}}^{d}$ denotes the space of all zero-one vectors of length $d$.
We write the $i$-th element of a vector $v$ as $v_i$ and the $(i,j)$-entry of a matrix $X$ as $X_{ij}$.

For a vector $v\in\mathbb{R}^d$ and $0 < p < \infty$, let $\|v\|_p = {(\sum_{i=1}^d |v_i|^p)}^{\frac{1}{p}}$ be the $l_p$-norm, and $\|v\|_\infty = \max_i |v_i|$ be the $l_\infty$-norm.
For a matrix $M\in\mathbb{R}^{d_1\times d_2}$, let $\|M\|_{\mathrm{F}} = \sqrt{\sum_{(i,j)\in [d_1]\times[d_2]} M_{ij}^2}$ and $\|M\|_{\infty} = \max_{(i,j)\in [d_1]\times[d_2]} M_{ij}$ be the Frobenius norm and the entry-wise infinity norm, respectively.
With two norms $l_p$ and $l_q$ on $\mathbb{R}^{d_2}$ and $\mathbb{R}^{d_1}$ respectively, define an operator norm of $M$ as $\|M\|_{p,q} = \sup_{\|x\|_p=1} \|Mx\|_q$.

For the sake of simplicity, for a set of matrix indices $\Omega\subseteq[d_1]\times[d_2]$, we write $\sum_{(i,j)\in\Omega}$ as $\sum_{\Omega}$, as long as the meaning is clear from the context.
Throughout this paper, we consider matrices of size $d_1\times d_2$.

\subsection{Quantization process and observation process}
A matrix completion is the problem of recovering an underlying target matrix $M\in\mathbb{R}^{d_1\times d_2}$ given its partial information.
Here, we define how this input is generated from $M$ in the binary matrix completion (BMC) problem.
Basically, we follow the definition of \citet{Daven14}.

In the BMC problem, a set of observed indices $\Omega = {\{(i_t,j_t)\}}_{t=1}^n\subset[d_1]\times[d_2]$ and corresponding entries of a matrix $Y\in{\{\pm 1\}}^{d_1\times d_2}$ called the quantization matrix are given as an input.
We call a matrix $A\in{\{-1,0,+1\}}^{d_1\times d_2}$ defined as follows the observation matrix.
\begin{align}
  A_{ij} = \begin{cases}
    Y_{ij} & \text{if } (i,j) \in \Omega,\\
    0 & \text{otherwise}.
  \end{cases}
\end{align}
The generation process of the input consists of two steps, how the values of M are quantized (i.e., how $Y$ is generated) and which entries are observed (i.e., how $\Omega$ is chosen).
We assume that these two steps are independent and we can consider them separately.
That is, the generation processes of $Y$ and $\Omega$ do not depend on each other.
Below, we explain these two steps in more detail.

\subsubsection{Quantization process}
We first consider the former step.
In a standard matrix completion setting with exact observations, we get a true value of the underlying matrix, i.e., $Y = M$.
When noisy observations are considered, we are given noisy values instead of exact ones, that is, $Y = M + Z$ where $Z\in\mathbb{R}^{d_1\times d_2}$ is a matrix containing noise.
We assume these noisy elements ${\{Z_{ij}\}}_{(i,j)\in[d_1]\times [d_2]}$ are independent and identically distributed (i.i.d.) according to a fixed distribution.

On the other hand, in the binary matrix completion setting, observed values are quantized into a binary value $\pm 1$.
First consider the following thresholding model.
\begin{align}\label{eq:threshold_model}
  Y_{ij} = \begin{cases}
    +1 & \text{if } M_{ij} + Z_{ij} \ge 0,\\
    -1 & \text{if } M_{ij} + Z_{ij} < 0.
  \end{cases}
\end{align}
Here for each entry $(i,j)$, a noisy value $M_{ij}+Z_{ij}$ is generated in the same way as the noisy setting, then it is quantized into $\pm 1$ according to the threshold value $0$.
Since there are no assumptions on the noise matrix $Z$ we can use any value as the threshold without loss of generality, and we simply use zero.

Interestingly, the noise term $Z$ plays an important role in the well-posedness of the binary matrix completion problem~\citep{Daven14}.
That is to say, if we set $Z$ to be constant, the problem becomes ill-posed.
To see this, consider the case where the target matrix $M$ can be decomposed as $M=uv^\top$ with some vectors $u\in\mathbb{R}^{d_1}$, $v\in\mathbb{R}^{d_2}$, and $Z=0$.
Then we can easily see that replacing an element of $u$ or $v$ with a value of the same sign yields the same quantization matrix.
This means even if we observe all entries of $Y$, we cannot distinguish $M$.
This ill-posedness does not change if we have further information of $M$, such as the norm.
However, when the stochastic noise $Z$ is introduced, the problem becomes well-posed, and we can recover $M$ in some degree of accuracy similarly to the standard setting~\citep{Daven14}.

Next, to make this model more tractable, we transform it and remove $Z$.
Letting the cumulative density function (CDF) of the distribution of the negative noise $-Z$ as $f$, the above model can be rewritten as follows.
\begin{align}\label{eq:prob_model}
  Y_{ij} = \begin{cases}
    +1 & \text{with probability } f(M_{ij}),\\
    -1 & \text{with probability } 1 - f(M_{ij}).
  \end{cases}
\end{align}
We call this $f$ a quantization probability function (QPF).
As long as the QPF satisfies properties of the CDF of some distribution, this model is equivalent to the previous one with that distribution.
For the purpose of theoretical analysis, we make some more assumptions on the QPF.
For $\alpha > 0$, we define two quantities $L_{\alpha}$ and $\beta_{\alpha}$ as follows.
\begin{align}\label{def:l_and_beta}
  L_{\alpha} \defeq \sup_{|x|\le\alpha}\frac{|f'(x)|}{f(x)(1-f(x))}, \quad
  \beta_{\alpha} \defeq \sup_{|x|\le\alpha}\frac{f(x)(1-f(x))}{{(f'(x))}^2}.
\end{align}
We assume that these terms are well-defined with $f$,
more precisely, $f$ is differentiable and takes a value in range $(0,1)$
for $x\in[-\alpha,\alpha]$, and $f'$ is non-zero in $[-\alpha,\alpha]$.

We list some possible choices for the QPF, proposed in \citet{Daven14}.
\begin{itemize}
\item{
  Probit regression / Gaussian noise\\
  The probit regression model is represented by the model~\eqref{eq:prob_model} with $f(x) = \Phi(\frac{x}{\sigma})$, where $\Phi$ is the CDF of Gaussian distribution $N(0,1)$, $N(a,b^2)$ denotes the Gaussian distribution with mean $a$ and variance $b^2$, and $\sigma$ is a parameter for the standard deviation.
  This is equivalent to the model~\eqref{eq:threshold_model}, where $Z_{ij}$ are i.i.d.\ according to $N(0,\sigma^2)$.
  We have
  \begin{align}\label{def:l_and_beta_gauss}
    L_{\alpha}\le\frac{4}{\sigma}(\frac{\alpha}{\sigma}+1),\quad \text{and} \quad
    \beta_{\alpha}\le\pi\sigma^2\exp(\frac{\alpha^2}{2\sigma^2}).
  \end{align}
}
\item{
  Logistic regression / Logistic noise\\
  The logistic regression model is represented by the model~\eqref{eq:prob_model} with $f(x) = \frac{1}{1 + \exp(-x)}$, or equivalently the model~\eqref{eq:threshold_model} where $Z_{ij}$ are i.i.d.\ according to the standard logistic distribution.
  We have
  \begin{align}\label{def:l_and_beta_log}
    L_{\alpha}=1,\quad \text{and} \quad
    \beta_{\alpha}=\frac{{(1+e^{\alpha})}^2}{e^{\alpha}}.
  \end{align}
}
\end{itemize}

\subsubsection{Observation process}\label{sec:obs_process}
Next we consider how the set of observed indices $\Omega$ is chosen.
We define three models used in existing methods \citep{Candes09, Cai13, Daven14}.
Let $\Pi = {\{\pi_{ij}\}}_{(i,j)\in[d_1]\times[d_2]}$ be a distribution over $[d_1]\times[d_2]$, which satisfies $\pi_{ij}\in[0,1]$ and $\sum_{[d_1]\times[d_2]} \pi_{ij} = 1$.
We consider the case where the number of observed indices is $n$.

One is a multi-Bernoulli model, which is used in \citet{Daven14}.
In this model, each entry $(i,j)$ is observed independently according to the distribution $\Pi$ rescaled so that $E[|\Omega|] = n$, that is, $P((i,j)\in\Omega) = n\pi_{ij}$ for all $(i,j)\in[d_1]\times[d_2]$, where $P(\cdot)$ denotes the probability of an event.
Note that the expected size of $\Omega$ is $n$, and in practice, we cannot know the actual $n$ used in the process.

The second is a multinomial model, which is used in \citet{Cai13}.
In this model, we repeatedly sample an index according to $\Pi$ with replacement $n$ times.
Formally, $\Omega={\{(i_t,j_t)\}}_{t=1}^n\in{([d_1]\times[d_2])}^n$ and
$P((i_t,j_t) = (k,l)) = \pi_{kl}$ for all $t\in[n]$ and $(k,l)\in[d_1]\times[d_2]$.
This model is analogous to the sampling model of the classification problem.
However, there exists a clear drawback that we sample some entries multiple times with high probability.

The last one is an \emph{all-at-once} model, which is used in \citet{Candes09}.
Given $\Pi$, the joint probability for the set of indices can be calculated.
This model directly samples $\Omega$ of size $n$ according to these probabilities.

The most popular choice for the sampling distribution $\Pi$ is the uniform observation assumption, where all of $\pi_{ij}$ have the same value $\frac{1}{d_1 d_2}$.
We denote this distribution by $\Pi_{\mathrm{uni}}$.
This simple assumption has an advantage in theoretical analysis and has been used in previous research such as \citet{Rong16}.
However, from the viewpoint of real-world application, the uniform observation assumption seems too idealistic and not to hold \citep{Cai16}, so we leave $\Pi$ general in this paper.

\subsection{Underlying matrix and constraints}\label{sec:constraint}
To recover the underlying matrix, we make some assumptions on $M$, since, without any assumptions, unobserved entries can take arbitrary values.
Here, we discuss what kind of assumptions we will use.

A basic assumption used in the matrix completion problem is the low-rankness of $M$, that is, for small $r \ll \min(d_1, d_2)$, $M$ satisfies $\rank(M) \le r$.
This is equivalent to that $M$ can be factorized as $UV^{\top} = M$, where $U \in \mathbb{R}^{d_1\times r}$ and $V \in \mathbb{R}^{d_2\times r}$.
From the viewpoint of singular values, the assumption $\rank(M) = r$ means that the first $r$ singular values of $M$ are non-zero and others are exactly zero.
However, in many real-world applications, the singular values of $M$
gradually decreases to zero~\citep{Daven14}.
Thus separating them into zeros and non-zeros exactly can be problematic.
Another problem of the rank constraint is that the optimization problem becomes non-convex and NP-hard in general~\citep{Fazel02}.
So in this paper, we use relaxation of the low-rank constraint.

A popular option is the nuclear norm (a.k.a.\ the trace norm) $\|\cdot\|_{*}$, which is the sum of singular values.
In contrast to the rank, which counts the number of non-zero singular values, the nuclear norm takes the sum of singular values.
As discussed in \citet{Fazel02}, as a function of a matrix, the nuclear norm is a convex envelope of the rank, and thus replacing the rank constraint in the optimization problem with the nuclear norm makes the problem convex.
This is a big advantage of using this constraint.

Another option is the max norm, which is defined as follows.
\begin{align}
  \|M\|_{\max} = \inf_{U,V: M=UV^\top} \{ \|U\|_{2,\infty} \|V\|_{2,\infty} \}.
\end{align}
The max norm is also a convex surrogate of the rank~\citep{Foygel11}, and comparable to the trace norm, which can be written as
\begin{align}
  \|M\|_{*}    = \inf_{U,V: M=UV^\top} \{ \|U\|_{\mathrm{F}} \|V\|_{\mathrm{F}} \}
  = \frac{1}{2} \inf_{U,V: M=UV^\top} \{ \|U\|_{\mathrm{F}}^2 + \|V\|_{\mathrm{F}}^2 \}.
\end{align}
From above and an inequation $\|M\|_{\mathrm{F}} \le \sqrt{\max(d_1,d_2)}
\|M\|_{2,\infty}$, we have
\begin{align}
  \|M\|_{*} \le \sqrt{d_1 d_2}\|M\|_{\max}.
\end{align}
For more discussions on the comparison of the trace and max norms, see
\citet{Cai13} and \citet{Srebro05rank}.

In addition to them, we also constrain the infinity norm.
When we use a QPF listed above, entries with a large absolute value will be quantized almost deterministic.
Thus this assumption is important to keep the randomness of the quantization process.
Overall, we focus on matrices in the following spaces.
\begin{align}
  K_{*}(\alpha, r)    &\defeq \left\{M\in\mathbb{R}^{d_1\times d_2} ~|~
  \|M\|_{\infty}\le\alpha, \|M\|_{*}\le\alpha\sqrt{r d_1 d_2} \right\},\label{def:k_trace}\\
  K_{\max}(\alpha, C) &\defeq \left\{M\in\mathbb{R}^{d_1\times d_2} ~|~
  \|M\|_{\infty}\le\alpha, \|M\|_{\max}\le C \right\}. \label{def:k_max}
\end{align}
Here $\alpha$, $r$, $C$ are free parameters to be determined.
For a matrix $M$ of rank $r$, we have
\begin{align}
  \|M\|_{\mathrm{F}} \le \|M\|_{*} \le \sqrt{r}\|M\|_{\mathrm{F}}
  \le \sqrt{r d_1 d_2}\|M\|_{\infty},\\
  \text{and}\quad \|M\|_{\infty} \le \|M\|_{\max} \le \sqrt{r}\|M\|_{1,\infty}
  \le \sqrt{r}\|M\|_{\infty}.
\end{align}
From these inequations, we can consider $K_{*}$ and $K_{\max}$ as relaxed
versions of the rank constraint and the infinity norm constraint.
As discussed, if a matrix $M$ satisfies $\rank(M)\le r$ and $\|M\|_{\infty}\le\alpha$, then
\begin{align}
  M \in K_{\max}(\alpha, \alpha \sqrt{r}) \subset K_{*}(\alpha, r).
\end{align}

\section{Existing methods}\label{ch:mc}
In this section, we review existing studies on the matrix completion problem.
Although studies on this problem have a long history and there are various kinds of problem settings and approaches, we mainly focus on studies which are directly related to our method and ones which give a theoretical recovery guarantee.

\subsection{Matrix completion}
First, we review the standard matrix completion problem, where quantization does not happen.
Let $M\in\mathbb{R}^{d_1\times d_2}$ be a target matrix, $A\in\mathbb{R}^{d_1\times d_2}$ be an input matrix which contains observed values, and $\Omega\subset [d_1]\times[d_2]$ be a set of observed indices.
The common problem setting assumes $\rank(M)\le r $ for some constant $r>0$~\citep{Jain15}, and then the optimization problem to solve becomes
\begin{align}\label{opt:min_disc_prob}
  \begin{split}
    \min_{X\in\mathbb{R}^{d_1\times d_2}} \quad & \|P_{\Omega}(X-A)\|_{\mathrm{F}}^2\\
    \st \quad & \rank(X) \le r,
  \end{split}
\end{align}
where $P_\Omega$ is a projection which sets each entries not in the set $\Omega$ to $0$.
Here $\|P_{\Omega}(\cdot)\|_{\mathrm{F}}^2$ measures the discrepancy between the estimation and the observation.
The following alternative form is used in \citet{Jiang17}.
\begin{align}\label{opt:min_rank_prob}
  \begin{split}
    \min_{X\in\mathbb{R}^{d_1\times d_2}} \quad & \rank(X)\\
    \st \quad & \|P_{\Omega}(X-A)\|_{\mathrm{F}}^2 \le t,
  \end{split}
\end{align}
where $t\ge 0$ is a constant.
As discussed in \citet{Fazel02}, these two problems are closely related and the solutions become equivalent by properly adjusting $r$ and $t$.

Because of the discreteness of the rank operator, these optimization problems are NP-hard in general~\citep{Fazel02}, and thus it is difficult to solve directly.
There are two major approaches to solve them efficiently.
One is based on the technique called matrix factorization or alternating minimization \citep{Koren09, Jain13}.
This method decomposes the estimation matrix as $X = UV^\top$ using matrices $U\in\mathbb{R}^{d_1\times k}$ and $V\in\mathbb{R}^{d_2\times k}$ for $k\le d_1 \wedge d_2$, and iteratively optimizes $U$ and $V$.

Another approach uses the relaxation of the rank constraint.
As the trace norm and the max norm are semi-definite representable \citep{Fazel02, Srebro05max}, replacing the rank constraint with them enables us to solve the problem using well-studied semi-definite programming.
We introduce some theoretical guarantees of recovery proved in former studies.

\subsubsection{Uniform sampling distribution with the trace norm}
For the all-at-once model with the uniform observation distribution, there are many types of research \citep{Daven16}.
Here we consider trace norm relaxation of the optimization problem in Eq.~\eqref{opt:min_disc_prob}.
In the specific case where we are given noise-free observations, \citet{Candes09} proved a strong guarantee that under some assumptions, we can recover the {\it exact} target matrix by solving the following optimization problem:
\begin{align}\label{opt:trace_exact}
  \begin{split}
    \min_{X\in\mathbb{R}^{d_1\times d_2}} \quad & \|X\|_{*}\\
    \st \quad & P_{\Omega}(X) = P_{\Omega}(M).
  \end{split}
\end{align}

Their result is further improved by \citet{Recht11}.

Next, we introduce the work of \citet{Candes10},
which considers the noisy observation.
Let the noisy observation $Y = M + Z$.
Supposing that $\|P_{\Omega}(Z)\|_{\mathrm{F}} \le \delta$
with a positive constant $\delta>0$, consider the following optimization
problem.
\begin{align}\label{opt:candes10}
  \begin{split}
    \min_{X\in\mathbb{R}^{d_1\times d_2}} \quad & \|X\|_{*}\\
    \st \quad & \|P_{\Omega}(X-Y)\|_{\mathrm{F}} \le \delta.
  \end{split}
\end{align}
Then under some more assumptions on $M$,
the solution $M^{*}$ of this problem satisfies
\begin{align}
  \|M-M^{*}\|_{\mathrm{F}} \le 2\delta\left(1 + 2\sqrt{\frac{(2+p)(d_1\wedge d_2)}{p}}\right),
\end{align}
where $p = \frac{n}{d_1 d_2}$ is the fraction of observed entries.

\subsubsection{General sampling distribution with the max norm}
After the work of \citet{Foygel11} which used the max norm constraint and the uniform observation model, \citet{Cai16} studied the case of general observation distributions.
They solved the following optimization problem:
\begin{align}\label{opt:cai16}
  \begin{split}
    \min_{X} \quad & \sum_{t=1}^n {(Y_{i_t j_t} - X_{i_t j_t})}^2\\
    \st \quad & X \in K_{\max}(\alpha,\alpha\sqrt{r}),
  \end{split}
\end{align}
where $K_{\max}$ is defined in Eq.~\eqref{def:k_max},
under the following assumptions.
\begin{itemize}
\item{
  The set of observed indices $\Omega$ is drawn according to
  the multinomial model defined in Sec.~\ref{sec:obs_process},
  with a general distribution $\Pi$.
}
\item{
  The given noisy observations ${\{Y_{i_t j_t}\}}_{t=1}^n$ indexed by $\Omega$ satisfy
  $Y_{i_t j_t} = M_{i_t j_t} + \sigma z_t$, where $\sigma$ is a constant
  and $z_t$ is i.i.d.\ noise with mean $0$ and variance $1$.
}
\item{
  For constants $r$ and $\alpha$, $M \in K_{\max}(\alpha, \alpha\sqrt{r})$.
  As discussed in Sec.~\ref{sec:constraint},
  this is relaxation of $\rank(M)\le r$ and $\|M\|_{\infty}\le\alpha$.
}
\end{itemize}

With the uniform sampling distribution, it is common to use the scaled Frobenius norm
$\frac{1}{d_1 d_2}\|\cdot\|_{\mathrm{F}}$ to measure the estimation error.
As sampling distribution $\Pi$ is arbitrary, we rescale it according to $\Pi$.
For a matrix $X$, define the weighted Frobenius norm as follows.
\begin{align}\label{def:weight_frob}
  \|X\|_{\Pi} \defeq \sqrt{E_{(i,j)\sim\Pi}\left[X_{ij}^2\right]}\\
  = \sqrt{\sum_{[d_1]\times[d_2]} \pi_{ij} X_{ij}^2}.
\end{align}
Note that when $\Pi = \Pi_\mathrm{uni}$,
this is equivalent to the scaled Frobenius norm.

Then the following theorem holds.
\begin{theorem}\label{thm:cai16} (\citet{Cai16})
  Suppose that $d_1 + d_2 < n < d_1 d_2$ and
  noise sequence ${\{z_t\}}_{t=1}^n$ are independent sub-exponential random variables.
  That is, there exists a constant $K>0$ such that
  \begin{align}
    \max_{t\in[n]} E\left[\exp\left(\frac{|z_t|}{K}\right)\right] \le e,
  \end{align}
  where $e$ is the Napier's constant.
  Then for the solution $M^{*}$ of the optimization problem
  in Eq.~\eqref{opt:cai16} with probability at least $1 - 2e^{-(d_1+d_2)}$,
  \begin{align}
    \|M^{*}-M\|_{\Pi}^2
    \le C (\alpha \vee K\sigma) \alpha\sqrt{\frac{r(d_1 + d_2)}{n}},
  \end{align}
  where $C$ is an absolute constant.
  If in addition, $\pi_{ij}\ge\frac{1}{\nu d_1 d_2}$ is satisfied
  for all $(i,j)\in[d_1]\times[d_2]$ and a constant $\nu\ge1$,
  with probability at least $1 - 2e^{-(d_1+d_2)}$,
  \begin{align}
    \frac{1}{d_1 d_2}\|M^{*}-M\|_{\mathrm{F}}^2
    \le C\nu (\alpha \vee K\sigma) \alpha\sqrt{\frac{r(d_1 + d_2)}{n}}.
  \end{align}
\end{theorem}

\subsection{Binary matrix completion}
The binary matrix completion (BMC) problem, proposed by \citet{Daven14},
aims to recover the underlying target matrix given binary quantized observations.
There are several papers which consider the quantized input before their work
such as \citet{Srebro05rank} and \citet{Srebro05max}.
However, they only focused on the classification task,
that is, the recovery of only signs of the target matrix,
while the BMC problem aims to recover the actual target matrix.

\subsubsection{Binary matrix completion with the trace norm constraint}\label{sec:daven}
The problem setting of \citet{Daven14} is basically same as that
we introduced in Sec.~\ref{ch:pre}.
Let $M$ be the target matrix, $\Omega$ be the set of observed indices,
$Y$ be the quantization matrix and $A$ be the observation matrix.
Specific assumptions used in their work are as follows.
\begin{itemize}
\item{
  For constants $r$ and $\alpha$, $M \in K_{*}(\alpha, r)$.
  $K_{*}$ is defined in Eq.~\eqref{def:k_trace}.
  As discussed in Sec.~\ref{sec:constraint},
  this is relaxation of $\rank(M)\le r$ and $\|M\|_{\infty}\le\alpha$.
}
\item{
  Both $L_{\alpha}$ and $\beta_{\alpha}$ (defined in Eq.~\eqref{def:l_and_beta}) are well-defined
  with the quantization probability function (QPF) $f$.
}
\item{
  In the observation process, $\Omega$ is drawn according to
  the multi-Bernoulli model with the uniform distribution
  $\Pi_\mathrm{uni}$, with $E[|\Omega|]=n$.
}
\end{itemize}

Under these assumptions, we can write the entire generation process of $A$ as follows.
\begin{align}
  A_{ij} = \begin{cases}
    +1 & \text{with probability } (1-\rho)f(M_{ij}),\\
    -1 & \text{with probability } (1-\rho)(1 - f(M_{ij})),\\
    0 & \text{with probability } \rho,
  \end{cases}
\end{align}
where $\rho = 1-\frac{n}{d_1 d_2}$ is a misobservation rate.
Based on these probabilities, we can derive the likelihood of each entry of the estimation.
The negative log-likelihood function for an entry and an entire matrix are defined as follows.
\begin{align}\label{def:loss_func}
  l(x,a) \defeq & -\mathbbm{1}_{a=+1}\log f(x) - \mathbbm{1}_{a=-1}\log (1-f(x)),\\
  L(X,A) \defeq & \sum_{[d_1]\times[d_2]} l(X_{ij}, A_{ij}).
\end{align}
Note that it is equivalent to define $L(X,A) = \sum_{\Omega} l(X_{ij}, A_{ij})$ in this setting, but we use this general definition for later use.
\citet{Daven14} used this $L$ to measure the discrepancy between the estimation and observation matrices.
The optimization problem is defined as follows.
\begin{align}\label{opt:bmc_trace}
  \begin{split}
    \min_{X} \quad & L(X,A)\\
    \st \quad & X \in K_{*}(\alpha, r).
  \end{split}
\end{align}

For the solution $M^{*}$ of this problem,
they obtained an upper bound of the estimation error.
\begin{theorem} (\citet{Daven14})
  Under the above assumptions, with probability at least $1-\frac{C_1}{d_1+d_2}$,
  \begin{align}
    \frac{1}{d_1d_2}\|M-M^{*}\|_{\mathrm{F}}^2 \le
    C_\alpha \sqrt{\frac{r(d_1+d_2)}{n}}\sqrt{1+\frac{(d_1+d_2)\log(d_1 d_2)}{n}},
  \end{align}
  where $C_{\alpha}\defeq C_2 L_\alpha \beta_\alpha$
  and $C_1$ and $C_2$ are absolute constants.
  If $n\ge(d_1+d_2)\log(d_1 d_2)$ then this is simplified to
  \begin{align}
    \frac{1}{d_1d_2}\|M-M^{*}\|_{\mathrm{F}}^2 \le \sqrt{2} C_\alpha \sqrt{\frac{r(d_1+d_2)}{n}}.
  \end{align}
\end{theorem}

\subsubsection{Binary matrix completion with the max norm constraint}
As discussed in Sec.~\ref{sec:obs_process},
the uniform observation assumption used in \citet{Daven14}
is too idealistic for some practical applications.
Instead of this, \citet{Cai13} considered a more general observation model.
They also used the max norm constraint in exchange for the trace norm.
Assumptions used are:
\begin{itemize}
\item{
  For constants $r$ and $\alpha$, $M \in K_{\max}(\alpha, \alpha\sqrt{r})$.
  This is also relaxation of $\rank(M)\le r$ and $\|M\|_{\infty}\le\alpha$.
}
\item{
  Both $L_{\alpha}$ and $\beta_{\alpha}$ (defined in Eq.~\eqref{def:l_and_beta}) are well-defined
  with the QPF $f$.
}
\item{
  In the observation process, $\Omega$ is drawn according to the multinomial model with a general distribution $\Pi$.
}
\end{itemize}

The optimization problem is basically same as the one in Sec.~\ref{sec:daven} and only the constraint is changed.
\begin{align}\label{opt:bmc_max}
  \begin{split}
    \min_{X} \quad & L(X, A)\\
    \st \quad & X \in K_{\max}(\alpha, \alpha\sqrt{r}).
  \end{split}
\end{align}

We define a term $U_{\alpha}$ used in the estimation error upper bound for
the solution $M^{*}$ of this problem proved in \citet{Cai13}.
\begin{align}\label{def:u_alpha}
  U_{\alpha} \defeq \sup_{|x|\le\alpha}\log\frac{1}{f(x)(1-f(x))}.
\end{align}
This term is well-defined under the second assumption.
Then the following theorem holds.
\begin{theorem}\label{thm:cai13} (\citet{Cai13})
  Under the above assumptions, with probability at least $1-\delta$,
  \begin{align}
    \|M-M^{*}\|_{\Pi}^2 \le
    C \beta_{\alpha} \left(
    L_{\alpha}\alpha\sqrt{\frac{r(d_1+d_2)}{n}}
    + U_{\alpha}\sqrt{\frac{\log(4/\delta)}{n}}
    \right),
  \end{align}
  where $C$ is an absolute constant.
\end{theorem}

This bound is comparable to the one shown in Th.~\ref{thm:cai16},
\begin{align}
  \|M-M^{*}\|_{\Pi}^2 \le
  C' (\alpha \vee K\sigma) \alpha \sqrt{\frac{r(d_1 + d_2)}{n}}.
\end{align}
Consider the special case where the noise ${\{z_t\}}_{t=1}^n$ is taken from
the Gaussian distribution $\mathcal{N}(0,\sigma^2)$,
or equivalently the QPF is $f(x) = \Phi(x/\sigma)$.
Then we have
\begin{align}
  L_{\alpha}\le\frac{4}{\sigma}\left(\frac{\alpha}{\sigma}+1\right),
  \beta_{\alpha}\le\pi\sigma^2\exp\left(\frac{\alpha^2}{2\sigma^2}\right),
  U_{\alpha}\le{\left(\frac{\alpha}{\sigma}+1\right)}^2,
\end{align}
and thus the bound in Th.~\ref{thm:cai13} becomes
\begin{align}
  \|M-M^{*}\|_{\Pi}^2 \le
  C \exp\left(\frac{\alpha^2}{2\sigma^2}\right)
  \left\{(\alpha+\sigma)\alpha\sqrt{\frac{r(d_1+d_2)}{n}}
  + {(\alpha+\sigma)}^2\sqrt{\frac{\log(4/\delta)}{n}}\right\}.
\end{align}
With inequality $(\alpha\vee\sigma)\le(\alpha+\sigma)\le2(\alpha\vee\sigma)$, they claim that there is no essential loss of recovery accuracy caused by the quantization of the observation, if $\frac{\alpha}{\sigma}$ is bounded.
On the other hand, if signal-to-noise ratio $\frac{\alpha}{\sigma}$ is large, that is, $\alpha \gg \sigma$, the setting becomes relatively noise-less and this bound deteriorates.

\subsection{PU matrix completion}\label{sec:pumc}
The PU matrix completion problem, proposed by \citet{Hsieh15},
is a further extension of the binary matrix completion setting.
This problem is named after the PU learning in the classification field,
which stands for ``positive and unlabeled'' \citep{Letou00}.
In this setting, we can only observe positive entries, i.e., entries quantized into $+1$.
Recommender systems and SNS link prediction
where only ``like'' and ``friendship'' are observed are possible applications.

\subsubsection{Problem setting}
The main change of the PU matrix completion setting from the binary matrix completion lies in the observation process.
In this setting, instead of observing a subset of whole entries, we only observe a subset of positively quantized entries.
Note that thus the observation process and the quantization process depend on each other.

Let $M\in\mathbb{R}^{d_1\times d_2}$ be the target matrix
and $Y\in{\{\pm 1\}}^{d_1\times d_2}$ be the quantization matrix.
$Y$ is generated in the same way as the binary matrix completion setting with the QPF $f$.
Given a misobservation rate $\rho\in(0,1)$,
the observation matrix $A\in{\{0,+1\}}^{d_1\times d_2}$ is observed
according to the following probabilities.
\begin{align}
  \begin{split}
    P(A_{ij} = +1) &= \rho\mathbbm{1}_{Y_{ij}=+1},\\
    P(A_{ij} = 0)  &= 1 - P(A_{ij} = +1),
  \end{split}
\end{align}
where where $\mathbbm{1}_c$ denotes the indicator function outputting $1$
if the condition $c$ is true and $0$ otherwise.
Using the definition of the quantization process,
we can rewrite this as
\begin{align}
  A_{ij} = \begin{cases}
    +1 & \text{with probability } (1-\rho)f(M_{ij}),\\
    0 & \text{otherwise}.
  \end{cases}
\end{align}
This model can also be considered as the multi-Bernoulli model
over only positive entries.

\subsubsection{Method}
It is easy to see that methods for the standard and binary matrix completion problems
are not applicable to this PU setting,
since a matrix whose entries are filled by $1$ can be a solution.
To overcome this problem, \citet{Hsieh15} introduced the idea of the classification problem.

Given a quantization matrix $Y$, consider an entry $(i,j)$ as an instance
and the corresponding quantized value $Y_{ij}$ as its label.
From this interpretation, we can regard $Y$ as a set of $d_1 d_2$ i.i.d.\ training samples
drawn from a distribution parametrized by
the target matrix $M$ and the QPF $f$.
Then our goal now is to estimate unknown parameters of the underlying distribution.
Since $Y$ has only two kinds of values, $\pm 1$,
we can consider this as a binary classification problem.

The actual input we are given is an observation matrix $A$,
where some entries of $Y$ are misobserved and become $0$.
However, the number of values in $A$ is still only two, that is, $0$ and $+1$.
Thus regarding $0$ in $A$ as negative labels,
we can consider what happened in the observation process as
a label perturbation or the addition of label noise.
Precisely, in the observation process,
positive labels are flipped to negative labels with a probability $\rho$
while negative labels are not.

\citet{Natara13} studied the classification problem with this kind of label perturbation.
Let $\rho_{+1}$ and $\rho_{-1}$ be the perturbation rates of
the positive and negative labels respectively, that is,
the true label $y\in\{\pm 1\}$ and the corresponding perturbed label
$\tilde{y}\in\{\pm 1\}$ satisfy
\begin{align}
  \begin{split}
    P(\tilde{y} = -1 | y = +1) = \rho_{+1},\\
    P(\tilde{y} = +1 | y = -1) = \rho_{-1}.
  \end{split}
\end{align}
Then the following theorem tells us how to construct
an unbiased estimator of the loss with noisy labels.
\begin{theorem}\label{thm:natara13} (\citet{Natara13})
  Given any loss function $l\colon\mathbb{R}\times\{\pm 1\}\to\mathbb{R}$,
  define $\tilde{l}$ as
  \begin{align}
    \tilde{l}(x,y) \defeq \frac{(1-\rho_{-y})l(x,y) - \rho_{y}l(x,-y)} {1-\rho_{+}-\rho_{-}}.
  \end{align}
  Then for any $x$ and $y$, we have
  \begin{align}
    E_{\tilde{y}}[\tilde{l}(x,\tilde{y})] = l(x,y).
  \end{align}
\end{theorem}

\citet{Hsieh15} used this modified loss function in their method.
Since in the PU matrix completion setting, $\rho_{+1} = \rho$ and $\rho_{-1} = 0$,
the modified loss function for given $l$ becomes
\begin{align}\label{def:pu_tilde_l}
  \tilde{l}(x,a) \defeq & \begin{cases}
    \frac{l(x,+1) -\rho l(x,-1)}{1-\rho} & (a = +1),\\
    l(x,-1) & (a = 0).
  \end{cases}
\end{align}
Note that negative labels are represented in different ways in the observation matrix $A$ and the quantized matrix $Y$ and we bridge this gap in this definition.
Define the loss for the entire matrix as follows.
\begin{align}\label{def:pu_loss_func}
  L_{\mathrm{PU}}(X, A) \defeq \sum_{[d_1]\times[d_2]} \tilde{l}(X_{ij}, A_{ij}).
\end{align}
Then from Th.~\ref{thm:natara13}, we have
\begin{align}
  E_{A}\left[L_{\mathrm{PU}}(X, A)\right] = E_{Y}\left[L(X,Y)\right].
\end{align}
Here $E_{A}$ is a expectation over the generation process of $Y$ and $\Omega$.
The optimization problem is given as follows.
\begin{align}\label{opt:pu_gene}
  \begin{split}
    \min_{X} \quad & L_{\mathrm{PU}}(X, A)\\
    \st \quad & X \in K_{*}(\alpha, \alpha\sqrt{r}).
  \end{split}
\end{align}
Here we leave the loss function $l$ free for the purpose of generality,
while their theoretical analysis focused on specific choices.

\subsubsection{Theoretical analysis}
\citet{Hsieh15} established estimation error bounds for their method with two specific choices of parameters.
Since the QPF used in one of them is not continuous and not comparable to our setting, we show the bound for another one only.
\begin{theorem} (\citet{Hsieh15})
  Suppose for a constant $t$, the target matrix satisfies
  $\|M\|_{*}\le t$ and $M_{ij}\in[0,1]$ for all $(i,j)\in[d_1]\times[d_2]$,
  and the QPF $f$ is an identity function $f(x) = x$.
  For the solution $M^{*}$ of the problem in Eq.~\eqref{opt:pu_gene}
  with the loss function $l(x,a) = {(x-a)}^2$, with probability at least $1-\delta$,
  \begin{align}
    \frac{1}{d_1 d_2}\|M^{*}-M\|_{\mathrm{F}}^2 \le
    2Ct\frac{\sqrt{d_1}+\sqrt{d_2}+\sqrt[4]{s}}{(1-\rho)d_1 d_2}
    + 6\frac{\sqrt{\log(2\delta)}}{(1-\rho)\sqrt{d_1 d_2}},
  \end{align}
  where $s$ is the number of $1$s in $Y$ and $C$ is a constant.
\end{theorem}

Consider the case where $t=\sqrt{r d_1 d_2}$ and $d_1 = d_2 = d$.
Introducing parameters used in the PU setting, we can confirm that the bound of \citet{Daven14} becomes essentially
$O\left(\frac{\sqrt{r}}{\sqrt{(1-\rho)d}}\right)$, and that of \citet{Hsieh15} is $O\left(\frac{\sqrt{r}}{(1-\rho)\sqrt{d}}\right)$.
So the loss of accuracy is only $\frac{1}{\sqrt{(1-\rho)}}$, although in the PU setting we do not have negative observations.

\section{PU learning in classification}\label{ch:class}
Our method is highly motivated by the result of \citet{Sakai17}.
In this section, we introduce the PU classification problem to explain their work.

\subsection{Classification problem}
The classification problem is one of the most fundamental problems in machine learning \citep{Mohri12,Shale14}.
Let $x\in\mathcal{X}$ be an input and $y\in\mathcal{Y}$ be its label, equipped with an underlying probability density $p(x,y)$.
The goal of the classification problem is to learn a function, which maps an input to its label, given a set of (usually) labeled samples.
We consider the binary classification problem, that is, $\mathcal{Y}=\{\pm 1\}$.

A major approach to this problem is the empirical risk minimization (ERM) \citep{Vapni95}.
For a loss function $l\colon\mathbb{R}\to\mathbb{R}$ which takes a small value with a large input, the risk $R$ of a function $f\colon\mathcal{X}\to\mathcal{Y}$ under the loss $l$ is defined as
\begin{align}\label{eq:clf-risk}
  R(f) = E_{(x,y)\sim p(x,y)}\left[l(y f(x))\right].
\end{align}
This risk measures the average prediction quality of $f$ under loss function $l$.
The optimal classifier $f^{*}$, called the Bayes classifier, is given by
\begin{align}
  f^{*} = \argmin_{f} R(f)
\end{align}
with the zero-one loss $l_{0\text{-}1}(x) = \mathbbm{1}_{x \le 0}$.
Since we do not know the underlying density $p(x,y)$, we cannot compute the risk, and it is also not possible to obtain the Bayes classifier directly.

In practice, we approximate the risk by given samples.
Let
\begin{align}
  S = {\{(x_i,y_i)\}}_{i=1}^n \overset{\text{i.i.d.}}{\sim} p(x,y)
\end{align}
be a training set.
In the ERM, instead of minimizing the above risk in Eq.~\eqref{eq:clf-risk}, we use its approximation $\widehat{R}$, called the empirical risk, defined by the average loss on the given training set.
\begin{align}
  \widehat{R} = \frac{1}{n} \sum_{(x,y)\in S} l(y f(x)).
\end{align}

Since optimizing the risk function with the zero-one loss is known as the NP-hard problem due to the discreteness of that loss~\citep{Nguyen13}, we use its surrogate to solve the optimization problem efficiently.
It is known that if there are an infinite number of training samples, the minimizer of $\widehat{R}$ with convex surrogate loss functions such as the squared loss function $l(x)={(1-x)}^2$ agrees with the minimizer of $\widehat{R}$ with zero-one loss \citep{Rosasco04}.

\subsection{PU classification}
The PU classification is a problem to learn a classifier from only positive and unlabeled samples \citep{Elkan08}.
This problem setting is conceivable in various applications.
\begin{itemize}
\item{
  The internet advertising, where positive (user's interest) samples are easy to collect via ``clicked'' data but negative samples are not identifiable from ``unclicked'' data since users frequently skip advertisements.
}
\item{
  The land-cover classification \citep{Li11}, where positive samples (building areas) are easy to label but negative samples (other areas) are too diverse to properly label.
}
\end{itemize}

\subsubsection{Two settings in PU classification}
As mentioned by \citet{Niu16}, there are two formulations for PU classification, called the one-sample (OS) and two-samples (TS) settings.
In the OS setting, first a set of samples $(x,y,s)$ are taken from the underlying density $p(x,y,s)$, where $s$ indicates whether $x$ is labeled ($s=1$) or not ($s=0$).
In PU classification, samples in the negative class are not labeled, i.e., $P(s=1 \mid y=-1)=0$, and each sample in the positive class is labeled with a probability $P(s=1\mid y=+1)=1-\rho$ for a constant $\rho\in[0,1]$.
\citet{Elkan08} showed that if $\rho$ is given, we can calculate the expectation over $p(x,y)$ of any function, and proposed several methods to estimate this $\rho$.
The method of \citet{Natara13}, which is used in the PU matrix completion method, is applicable to this OS setting, considering the unlabeled samples as negative and setting $\rho_{+1} = \rho$ and $\rho_{-1} = 0$.

On the other hand, in the TS setting, positive samples are taken from the density $p(x | y=+1)$ and unlabeled samples are taken from the marginal density $p(x)$. 
Thus contrarily to the OS setting, positive samples and unlabeled samples are independent.
\citet{Du14} proposed an ERM-based PU classification method and analyzed its theoretical properties.
Moreover, \citet{Niu16} theoretically compared the PU learning method with supervised learning methods and revealed that in some case, the performance of the PU learning method is superior to that of supervised learning methods even though PU learning cannot access labels for the negative class.
These works \citep{Du14, Niu16} led to a semi-supervised classification method \citep{Sakai17}, which can utilize unlabeled samples for training a classifier without strong assumptions for the data distribution, unlike existing methods that rely on the cluster assumptions \citep{Chape06}.
Since this paper is highly motivated by \citet{Sakai17}, we here review the ERM-based PU classification method.

\subsubsection{ERM-based PU classification}
Let us denote the sets of positive and unlabeled samples by
\begin{align}
  S_{\mathrm{P}}&\defeq{\{x_i\}}_{i=1}^{n_{\mathrm{P}}}
  \overset{\text{i.i.d.}}{\sim} p(x\mid y=+1), \\
  S_{\mathrm{U}}&\defeq{\{x_i\}}_{i=1}^{n_{\mathrm{U}}}
  \overset{\text{i.i.d.}}{\sim}
  p(x)=\theta p(x\mid y=+1) + (1-\theta)p(x\mid y=-1),
\end{align}
where $\theta$ is the class-prior $p(y=+1)$.
Let $E_{\mathrm{P}}$, $E_{\mathrm{N}}$, and $E_{\mathrm{U}}$ be the expectations over $p(x|y=+1)$, $p(x|y=-1)$, and $p(x)$, respectively.
Moreover, let us define
\begin{align}
  R_{\mathrm{P}}(f) = E_{\mathrm{P}}\left[l(+f(x))\right],\\
  R_{\mathrm{N}}(f) = E_{\mathrm{N}}\left[l(-f(x))\right],\\
  R_{\mathrm{U,P}}(f) = E_{\mathrm{U}}\left[l(+f(x))\right],\\
  R_{\mathrm{U,N}}(f) = E_{\mathrm{U}}\left[l(-f(x))\right].
\end{align}
Then, the risk of the supervised learning in Eq.~\eqref{eq:clf-risk} can also be expressed as
\begin{align}\label{eq:clf-pn-risk}
  R(f) &= E_{(x,y)\sim p(x,y)}\left[l(y f(x))\right]\\
  &= \theta E_{\mathrm{P}}\left[l(+f(x))\right]
  + (1-\theta) E_{\mathrm{N}}\left[l(-f(x))\right]\\
  &= \theta R_{\mathrm{P}}(f) + (1-\theta)R_{\mathrm{N}}(f).
\end{align}
We refer to this risk as the positive-negative risk (the PN risk) and denote it by $R_{\mathrm{PN}}(f)$.

In the TS setting, the following transformation enables us to calculate the risk in Eq.~\eqref{eq:clf-risk} using only $S_{\mathrm{P}}$ and $S_{\mathrm{U}}$~\citep{Du14}.
From the definition of the marginal density, we have
\begin{align}
  E_{\mathrm{U}}[l(-f(x))]
  = \theta E_{\mathrm{P}}[l(-f(x))] + (1-\theta)E_{\mathrm{N}}[l(-f(x))],
\end{align}
or equivalently
\begin{align}
  R_{\mathrm{U,N}}(f) = \theta E_{\mathrm{P}}[l(-f(x))] + (1-\theta)R_{\mathrm{N}}(f).
\end{align}
Then, plugging this into $R_{\mathrm{PN}}$, we obtain the risk in PU classification (the PU risk) by
\begin{align}
  R_{\mathrm{PN}}(f)
  &= \theta R_{\mathrm{P}}(f) + (1-\theta)R_{\mathrm{N}}(f)\\
  &= \theta E_{\mathrm{P}}\left[l(+f(x))\right]
  + \left\{ R_{\mathrm{U,N}}(f) - \theta E_{\mathrm{P}}[l(-f(x))] \right\}\\
  &= \theta \bar{R}_{\mathrm{P}}(f) + R_{\mathrm{U,N}}(f)\\
  &\eqdef R_{\mathrm{PU}}(f),\label{eq:clf-pu-risk}
\end{align}
where $\bar{R}_{\mathrm{P}}(f) \defeq E_{\mathrm{P}} \left[\bar{l}(f(x))\right]$ and $\bar{l}(t) \defeq l(t)-l(-t)$ is a composite loss.

In practice, we approximate the expectations in the PU risk by corresponding sample averages and obtain the empirical PU risk as
\begin{align}
  \widehat{R}_{\mathrm{PU}}(f) \defeq
  \frac{\theta}{n_{\mathrm{P}}}\sum_{x\in S_{\mathrm{P}}}\bar{l}(f(x))
  + \frac{1}{n_{\mathrm{U}}}\sum_{x\in S_{\mathrm{U}}}l(-f(x)).
\end{align}
By minimizing the empirical PU risk, we obtain a trained classifier from only positive and unlabeled samples.
Note that the class-prior $\theta=P(y=+1)$ is replaced with an estimate based on domain knowledge or some class-prior estimation method from PU data such as \citet{Christ15}.

\subsection{PNU Learning}
Based on the PU classification method in the TS setting, the semi-supervised classification method was proposed by \citet{Sakai17}.
The idea is to combine the risk in the supervised learning with the risk in the PU learning method, which enables us to approximate the risk using both labeled and unlabeled samples.

First, let us define negative samples as
\begin{align}
  S_{\mathrm{N}}\defeq{\{x_i\}}^{n_{\mathrm{N}}}_{i=1}
  \overset{\text{i.i.d.}}{\sim} p(x\mid y=-1).
\end{align}
Furthermore, we define the risk in classification from negative and unlabeled data (the NU risk) and the empirical risk as
\begin{align}
  R_{\mathrm{NU}}(f) & \defeq (1-\theta)\bar{R}_{\mathrm{N}}(f) + R_{\mathrm{U,P}}(f) , \\
  \widehat{R}_{\mathrm{NU}}(f) & \defeq
  \frac{(1-\theta)}{n_{\mathrm{N}}}\sum_{x\in S_{\mathrm{N}}}\bar{l}(-f(x))
  + \frac{1}{n_{\mathrm{U}}}\sum_{x\in S_{\mathrm{U}}}l(f(x)) ,
\end{align}
where $\bar{R}_{\mathrm{N}} \defeq E_{\mathrm{N}}[\bar{l}(-f(x))]$.
The NU classification is just a counterpart of the PU classification and the NU risk can be obtained in a similar way that we obtain the PU risk.

Then, the PNU risk is defined by
\begin{align}
  R_{\mathrm{PNU}}^{\eta}(f) \defeq
  \begin{cases}
    (1-\eta)R_{\mathrm{PN}}(f) + \eta R_{\mathrm{PU}}(f) & (\eta\geq0) , \\
    (1+\eta)R_{\mathrm{PN}}(f) - \eta R_{\mathrm{NU}}(f) & (\eta<0) ,
  \end{cases}
\end{align}
where $\eta\in[-1,+1]$ is the combination parameter.
The method minimizing this PNU risk is referred to as the PNU learning.
The PNU risk function is either convex combinations of the PN and PU risks or that of the PN and NU risks.
\citet{Sakai17} also considered the combinations of the PU and NU risks, but they revealed that the PNU risk is more promising from both theoretical and empirical viewpoints.

In practice, we use the empirical PNU risk.
\begin{align}
  \widehat{R}_{\mathrm{PNU}}^{\eta}(f) \defeq
  \begin{cases}
    (1-\eta)\widehat{R}_{\mathrm{PN}}(f) + \eta \widehat{R}_{\mathrm{PU}}(f) & (\eta\geq0) , \\
    (1+\eta)\widehat{R}_{\mathrm{PN}}(f) - \eta \widehat{R}_{\mathrm{NU}}(f) & (\eta<0) .
  \end{cases}
\end{align}
The combination parameter $\eta$ is determined by, e.g., cross-validation.

PNU learning is demonstrated to achieve higher classification accuracy than other existing methods in experiments \citep{Sakai17}.
This motivates us to consider a novel matrix completion approach based on PU matrix completion.

\subsection{Comparison to Matrix Completion}\label{sec:diff_cl_mc}
We propose using the idea of the PNU classification method, that is, the combination of PN and PU risks, in the BMC problem.
However, there are several differences between classification and matrix completion.
Here we list some of them.

\begin{itemize}
\item{Unlabeled samples:\\
  In the PU classification problem, we draw unlabeled samples from marginal density $p(x)$, while in the BMC problem, there are no unlabeled data.
  \citet{Hsieh15} treated unobserved entries as unlabeled data to address the PU setting.
  Note that in this case observed and unobserved entries are not independent and thus this corresponds to the OS setting.
  We follow this and regard unobserved entries as unlabeled.
}
\item{Fixed data size:\\
  In the classification problem, the number of given training samples can be arbitrary, while in the matrix completion, regarding each entry as a sample, there are just $d_1 d_2$ samples in total.
  This does not change the optimization process, but a theoretical analysis would be affected.
}
\item{What to estimate:\\
  In the classification problem, what we want to estimate is a function which maps a sample to its label.
  However, in the binary matrix completion, our goal is to estimate the underlying matrix, which rather corresponds to the parameters of the data distribution.
  Actually this is more difficult than just learning a classifier $g\colon[d_1]\times[d_2]\to{\{\pm 1\}}$, since we can construct this $g$ from the estimated matrix $M$ as $g((i,j)) = 2\mathbbm{1}_{f(M_{ij}) \ge \frac{1}{2}} -1$, but the reverse is impossible.
}
\end{itemize}

\section{Proposed method}\label{ch:proposed}
In this section, we discuss a way to improve former BMC methods and propose a new method.

\subsection{Motivation}
Again, let $M$, $Y$, and $A$ be the target, quantization and observation matrices, respectively.
Let $f$ be the QPF, $\Omega$ be a set of observed indices and $\rho$ be the misobservation rate.
First we reprint loss functions and risks used in \citet{Daven14} and \citet{Hsieh15}.
In the former, they used the negative log-likelihood function.
\begin{align}
  l_{\mathrm{NLL}}(x,a) &\defeq
  \mathbbm{1}_{a=+1} \log\frac{1}{f(x)} + \mathbbm{1}_{a=-1} \log\frac{1}{1-f(x)},\\
  L_{\mathrm{obs}}(X, A) &\defeq \sum_{[d_1]\times[d_2]} l_{\mathrm{NLL}}(X_{ij}, A_{ij}).
\end{align}
In the latter, for a given loss function $l$, they used a modified loss $\tilde{l}$.
\begin{align}
  \tilde{l}(x,a) &\defeq \begin{cases}
    \frac{1}{1-\rho}(l(x,+1) -\rho l(x,-1)) & (a = +1),\\
    l(x,-1) & (a = 0),
  \end{cases}\\
  L_{\mathrm{all}}(X, A) &\defeq \sum_{[d_1]\times[d_2]} \tilde{l}(X_{ij}, A_{ij}).
\end{align}
Hereafter we consider the case of $l = l_{\mathrm{NLL}}$.
Note that $A$ belongs to different spaces in each study, that is, the former supposes $A\in{\{-1,0,+1\}}^{d_1\times d_2}$ and the latter supposes $A\in{\{0,+1\}}^{d_1\times d_2}$.

The point we focus on is the set of indices which risks are defined over.
From the definition of $l_{\mathrm{NLL}}$, $L_{\mathrm{obs}}$ can be equivalently expressed as
\begin{align}
  L_{\mathrm{obs}}(X, A) \defeq \sum_{\Omega} l_{\mathrm{NLL}}(X_{ij}, A_{ij}),
\end{align}
for the BMC setting.
That is, $L_{\mathrm{obs}}$ is effectively defined only over observed entries.
On the other hand, $L_{\mathrm{all}}$ is defined over all entries including unobserved ones.
Although this is natural since their method is based on the PU classification method of \citet{Natara13}, this difference gives us an important insight.
That is, we can extract information even from unobserved entries, in other words, the ``unobservedness'' has information too.

In the standard methods for the matrix completion problem and the binary matrix completion problem, in the same way as $L_{\mathrm{obs}}$, the risk for the estimation matrix is not defined over unobserved entries.
Also, the method of \citet{Hsieh15} is constructed for the PU setting and not applicable to the BMC setting without modifications.
As far as we know, in the BMC setting, there are no methods which define a risk overall entries and distinguish three kinds of entries.
This motivates us to build a method for BMC, which can handle all of the positive, negative, and unobserved entries.

The work of \citet{Sakai17} suggested a way to achieve this.
They argue that the combination of the PU and PN methods yields better performance in the semi-supervised classification field.
Regarding the method of \citet{Hsieh15} as PU and the methods of \citet{Daven14} and \citet{Cai13} as PN, we can do the same thing in the matrix completion field.
Of course, since the problem settings are different, we cannot directly obtain the same result.
We later discuss what kind of combinations is better for our problem.

\subsection{Modification of PU method}
The PU matrix completion method cannot be directly applied to the BMC setting since the observation matrix $A$ in BMC can contain $-1$.
Here, we discuss how to solve this.

The problem here is how to treat observed negative entries.
The simplest solution to this would be to ignore them, that is, assign $0$ for the loss as follows.
\begin{align}
  \tilde{l}^{\prime}(x,a) &\defeq \begin{cases}
    \frac{1}{1-\rho}(l(x,+1) -\rho l(x,-1)) & (a = +1),\\
    l(x,-1) & (a = 0),\\
    0 & (a = -1),
  \end{cases}\\
  L_{\mathrm{all}}^{\prime}(X, A) &\defeq \sum_{[d_1]\times[d_2]} \tilde{l}(X_{ij}, A_{ij}).
\end{align}
However this modification eliminates the main property of $\tilde{l}$, that is, for any $X$,
\begin{align}\label{eq:exp_tilde_l}
  E_{A}\left[L_{\mathrm{all}}(X,A)\right] = E_{Y}\left[L_{\mathrm{obs}}(X,Y)\right].
\end{align}

Under the multi-Bernoulli observation model with the uniform sampling distribution, each entry of $A$ satisfies
\begin{align}
  A_{ij} = \begin{cases}
    +1 & \text{with probability } (1-\rho)f(M_{ij}),\\
    -1 & \text{with probability } (1-\rho)(1 - f(M_{ij})),\\
    0 & \text{with probability } \rho,
  \end{cases}
\end{align}
in the BMC setting, and
\begin{align}
  A_{ij} = \begin{cases}
    +1 & \text{with probability } (1-\rho)f(M_{ij}),\\
    0 & \text{with probability } (1-\rho)(1-f(M_{ij})) + \rho,
  \end{cases}
\end{align}
in the PU setting.
Thus to keep Eq.~\eqref{eq:exp_tilde_l} satisfied, it is enough to treat negative entries in the same way as unobserved ones, in other words, ignore negative labels as follows.
\begin{align}
  \tilde{l}^{\prime\prime}(x,a) &\defeq \begin{cases}
    \frac{1}{1-\rho}(l(x,+1) -\rho l(x,-1)) & (a = +1),\\
    l(x,-1) & (a = 0),\\
    l(x,-1) & (a = -1),
  \end{cases}\\
  L_{\mathrm{all}}^{\prime\prime}(X, A) &\defeq
  \sum_{[d_1]\times[d_2]} \tilde{l}^{\prime\prime}(X_{ij}, A_{ij}).
\end{align}
This satisfies
\begin{align}\label{eq:exp_pu}
  E_{A}\left[L_{\mathrm{all}}^{\prime\prime}(X,A)\right] = E_{Y}\left[L_{\mathrm{obs}}(X,Y)\right],
\end{align}
for any $X$.
So this $L_{\mathrm{all}}^{\prime\prime}$ is a natural extension of the PU risk $L_{\mathrm{all}}$ to the BMC setting.

Hereafter, we only consider the BMC problem.
So for the sake of simplicity, we refer to $L_{\mathrm{obs}}$ and $L_{\mathrm{all}}^{\prime\prime}$ as $L_{\mathrm{PN}}$ and $L_{\mathrm{PU}}$, respectively, and $\tilde{l}^{\prime\prime}$ as $\tilde{l}_{\mathrm{PU}}$.
We also define the risk based on observed negative entries and unobserved entries (the NU risk), $L_{\mathrm{NU}}$, as follows.
\begin{align}
  \tilde{l}_{\mathrm{NU}}(x,a) &\defeq \begin{cases}
    l(x,+1) & (a = +1),\\
    l(x,+1) & (a = 0),\\
    \frac{1}{1-\rho}(l(x,-1) -\rho l(x,+1)) & (a = -1),
  \end{cases}\\
  L_{\mathrm{NU}}(X, A) &\defeq \sum_{[d_1]\times[d_2]} \tilde{l}_{\mathrm{NU}}(X_{ij}, A_{ij}).
\end{align}
This is a counterpart of $L_{\mathrm{PU}}$ and also satisfies
\begin{align}\label{eq:exp_nu}
  E_{A}\left[L_{\mathrm{NU}}(X,A)\right] = E_{Y}\left[L_{\mathrm{PN}}(X,Y)\right],
\end{align}
for any $X$.

\subsection{Combination of existing methods}
Now we have three loss functions which are based on the former methods, namely, $L_{\mathrm{PN}}$, $L_{\mathrm{PU}}$, and $L_{\mathrm{NU}}$.
We note that all of them do not fully utilize the given information.
$L_{\mathrm{PN}}$ does not take unobserved entries into account and $L_{\mathrm{PU}}$ and $L_{\mathrm{NU}}$ are defined over all entries but ignore negative and positive labels, respectively.
Here, we construct the risk which can exploit all kinds of positive, negative, and unobserved entries.

We define the PUNU risk $L_{\mathrm{PUNU}}$ and the PNU risk $L_{\mathrm{PNU}}$ as follows, combining these loss functions.
\begin{align}
  L_{\mathrm{PUNU}}^{\gamma}(X,A) &\defeq
  (1-\gamma) L_{\mathrm{PU}}(X,A) + \gamma L_{\mathrm{NU}}(X,A),\\
  L_{\mathrm{PNU}}^{\eta}(X,A) &\defeq \begin{cases}
    L_{\mathrm{PNPU}}^{ \eta} & (\eta \ge 0),\\
    L_{\mathrm{PNNU}}^{-\eta} & (\eta < 0),
  \end{cases}
\end{align}
where
\begin{align}
  L_{\mathrm{PNPU}}^{\gamma}(X,A) &\defeq
  (1-\gamma) L_{\mathrm{PN}}(X,A) + \gamma L_{\mathrm{PU}}(X,A),\\
  L_{\mathrm{PNNU}}^{\gamma}(X,A) &\defeq
  (1-\gamma) L_{\mathrm{PN}}(X,A) + \gamma L_{\mathrm{NU}}(X,A),
\end{align}
and $\gamma\in[0,1]$ and $\eta\in[-1,+1]$ are combination parameters.
In the classification field, the PUNU risk works
poorly compared to the PNU risk \citep{Sakai17}.
In the BMC problem, we also obtain a similar result shown in the next section.

Here, we consider a further extension of these risks.
An important observation is that
the PUNU risk $L_{\mathrm{PUNU}}^{\gamma}$ satisfies the following equation.
\begin{align}\label{eq:exp_punu}
  E_{A}\left[L_{\mathrm{PUNU}}^{\gamma}(X,A)\right]
  &= E_{A}\left[(1-\gamma) L_{\mathrm{PU}}(X,A) + \gamma L_{\mathrm{NU}}(X,A)\right]\\
  &= (1-\gamma)E_{A}\left[L_{\mathrm{PU}}(X,A)\right]
  + \gamma    E_{A}\left[L_{\mathrm{NU}}(X,A)\right]\\
  &= E_{Y}\left[L_{\mathrm{PN}}(X,Y)\right],
\end{align}
for all $\gamma\in[0,1]$.
That is, the expected value of the PUNU risk $L_{\mathrm{PUNU}}^{\gamma}(X,A)$ equals to that of the ordinary risk over entire quantization matrix $L_{\mathrm{PN}}(X,Y)$ and thus it keeps the main property of $L_{\mathrm{PU}}$ and $L_{\mathrm{NU}}$.
Moreover, from its definition, we can write down it as follows.
\begin{align}\label{eq:expand_punu}
  L_{\mathrm{PUNU}}^{\gamma}(X,A)
  &= (1-\gamma) L_{\mathrm{PU}}(X,A) + \gamma L_{\mathrm{NU}}(X,A)\\
  &= \sum_{[d_1]\times[d_2]} \left[
    (1-\gamma)\tilde{l}_{\mathrm{PU}}(X_{ij}, A_{ij})
    + \gamma  \tilde{l}_{\mathrm{NU}}(X_{ij}, A_{ij}) \right]\\
  &= \sum_{[d_1]\times[d_2]} \tilde{l}_{\mathrm{PUNU}}^{\gamma}(X_{ij}, A_{ij}),
\end{align}
where
\begin{align}\label{eq:l_punu}
  \tilde{l}_{\mathrm{PUNU}}^{\gamma}(x,a)
  &\defeq (1-\gamma)\tilde{l}_{\mathrm{PU}}(x,a) + \gamma\tilde{l}_{\mathrm{NU}}(x,a)\\
  &= \begin{cases}
        \displaystyle
    \frac{1-\gamma\rho}{1-\rho} l(x,+1) - (1-\gamma)\frac{\rho}{1-\rho} l(x,-1) & (a = +1),\\
    \displaystyle
    \gamma l(x,+1) + (1-\gamma)l(x,-1) & (a = 0),\\
    \displaystyle
    -\gamma\frac{\rho}{1-\rho} l(x,+1) + \frac{1-(1-\gamma)\rho}{1-\rho} l(x,-1) & (a = -1).
  \end{cases}
\end{align}
We can see that $L_{\mathrm{PUNU}}^{\gamma}$ assigns different losses on each kind of positive, negative, and unobserved entries.
This means that, in addition to the above property, $L_{\mathrm{PUNU}}^{\gamma}$ can handle all kinds of entries properly.
Consequently, we can regard $L_{\mathrm{PUNU}}^{\gamma}$ as an extension of the PU and NU risks.

As we mentioned, $L_{\mathrm{PNU}}^{\eta}$ works better also in the BMC problem.
However, based on the above discussion on $L_{\mathrm{PUNU}}^{\gamma}$,
we consider all the combinations, i.e., combining
$L_{\mathrm{PU}}$, $L_{\mathrm{NU}}$, and $L_{\mathrm{PN}}$.
We define such a risk as follows.
\begin{align}\label{def:tri_3var}
  L_{\mathrm{TRI}}^{\gamma_{\mathrm{PN}},\gamma_{\mathrm{PU}},\gamma_{\mathrm{NU}}}(X,A)
  &= \gamma_{\mathrm{PN}} L_{\mathrm{PN}}(X,A)
  + \gamma_{\mathrm{PU}} L_{\mathrm{PU}}(X,A)
  + \gamma_{\mathrm{NU}} L_{\mathrm{NU}}(X,A),
\end{align}
where
\begin{align}
& \gamma_{\mathrm{PN}}, \gamma_{\mathrm{PU}}, \gamma_{\mathrm{NU}} \in [0,1],\\
\text{and} \quad & \gamma_{\mathrm{PN}} + \gamma_{\mathrm{PU}} + \gamma_{\mathrm{NU}} = 1.
\end{align}
Thus we can regard
$L_{\mathrm{TRI}}^{\gamma_{\mathrm{PN}},\gamma_{\mathrm{PU}},\gamma_{\mathrm{NU}}}$
as a weighted average of $L_{\mathrm{PN}}$, $L_{\mathrm{PU}}$, and $L_{\mathrm{NU}}$.
We experimentally show the superiority of this risk in the next section.
For the sake of simplicity, we refer to $L_{\mathrm{TRI}}^{\gamma_{\mathrm{PN}},\gamma_{\mathrm{PU}},\gamma_{\mathrm{NU}}}$ as $L_{\mathrm{TRI}}$ hereafter.

\subsection{Algorithms}
The optimization problem we want to solve is,
\begin{align}\label{opt:tri}
  \begin{split}
    \min_{X} \quad & L_{\mathrm{TRI}}(X,A)\\
    \st \quad & X \in K_{\max}(\alpha, R),
  \end{split}
\end{align}
where $K_{\max}$ is defined in Eq.~\eqref{def:k_max} and $\alpha$ and $R$ are parameters to be determined.
Here we use the max norm constraint based on the empirical result of \citet{Cai13}, that the max norm is superior to the trace norm.
Here, we describe how to obtain a solution to this problem.

There are several methods to solve the matrix completion problem, such as the singular value decomposition based method \citep{Chatt15} and the Frank-Wolfe type algorithm \citep{Jaggi13}.
However, from the viewpoint of computational complexity, those methods can be slow in practice and may not be applicable to the large-scale matrices.
So we use the matrix factorization based method, following \citet{Cai13}.

First consider decomposing $X\in\mathbb{R}^{d_1\times d_2}$ as $X = UV^\top$, using $U\in\mathbb{R}^{d_1\times k}$, $V\in\mathbb{R}^{d_2\times k}$ and a constant $1\le k\le (d_1\vee d_2)$.
More formally, for fixed $1\le k\le (d_1\vee d_2)$, define
\begin{align}
  \mathcal{M}_k(R) \defeq \{(U,V) |
  U\in\mathbb{R}^{d_1\times k}, V\in\mathbb{R}^{d_2\times k},
  \max(\|U\|_{2,\infty}^2, \|V\|_{2,\infty}^2) \le R \}.
\end{align}
Then we can rewrite the problem in Eq.~\eqref{opt:tri} as
\begin{align}\label{opt:tri_mf}
  \begin{split}
    \min_{U,V} \quad & L_{\mathrm{TRI}}(UV^\top,A)\\
    \st \quad & (U,V) \in \mathcal{M}_k(R), \|UV^\top\|_{\infty} \le \alpha.
  \end{split}
\end{align}

We can solve the problem in Eq.~\eqref{opt:tri_mf} by iterating the following update.
Note that under the assumption that the QPF $f$ is differentiable, $L_{\mathrm{TRI}}(X, A)$ is differentiable with respect to the first argument.
Let $(U^t, V^t)$ be estimators at a step $t = 1,2,\ldots$.
We first perform gradient descent.
\begin{align}\label{alg:grad_des}
  \begin{split}
    U_{1}^{t}
    &= U^t - \tau  \nabla L_{\mathrm{TRI}}(U^t{(V^t)}^\top, A) V^t,\\
    V_{1}^{t}
    &= V^t - \tau {\nabla L_{\mathrm{TRI}}(U^t{(V^t)}^\top, A)}^\top U^t,
  \end{split}
\end{align}
where $\tau > 0$ is a step-size parameter.
Then we project $(U_{1}^{t}, V_{1}^{t})$ onto $\mathcal{M}_k(R)$.
This projection is carried out by rescaling each row of $U_1^t$ and $V_1^t$, whose $l_2$-norms exceed $R$ so that their norm become $R$.
We keep rows with norm smaller than $R$ unchanged.
Let the result of this projection be $(U_2^t, V_2^t)$.
Finally, we rescale $(U_2^t, V_2^t)$ so that $\|U_2^t {(V_2^t)}^\top\|_{\infty} \le \alpha$.
That is, if $\|U_2^t {(V_2^t)}^\top\|_{\infty} > \alpha$,
\begin{align}
  \begin{split}
    U^{t+1} &= \sqrt{\frac{\alpha}{\|U_2^t {(V_2^t)}^\top\|_{\infty}}} U_2^t,\\
    V^{t+1} &= \sqrt{\frac{\alpha}{\|U_2^t {(V_2^t)}^\top\|_{\infty}}} V_2^t,
  \end{split}
\end{align}
and otherwise $(U^{t+1}, V^{t+1}) = (U_2^t, V_2^t)$.

According to \citet{Burer03}, it is important to use sufficiently large $k$, at least larger than the actual rank of the target matrix, to obtain the global optimum.
Also as \citet{Cai13} discussed, we should not use too large $k$ since that makes the optimization problem unnecessarily complex.
So following those studies, we use the scheme that iteratively increases $k$ from a small number until the resulting $M = UV^T$ converges.

\section{Experiments}\label{ch:exp}
In this section, we conduct several experiments to numerically investigate the performance of our method.

\subsection{Illustration of PUNU and PNU risks}
First, we show how the PUNU and PNU risks behave in the binary matrix completion.
We generate the target matrix $M$ by $M = UV^\top$, where entries of matrices $U\in\mathbb{R}^{d_1\times r}$ and $V\in\mathbb{R}^{d_2\times r}$ are drawn i.i.d.\ from the uniform distribution over $[-1,+1]$.
Then $M$ is normalized so that $\|M\|_{\infty}\le\alpha$.
We generate the quantization matrix $Y$ using QPF $f(x) = \Phi(\frac{x}{\sigma})$.
Each entry of $Y$ is independently observed with probability $1-\rho$.

We set $d_1=d_2=100$, $r=10$, $\alpha=1$, $\sigma=0.1$ and $\rho=0.85$ in this experiment, and assume that all of these parameters are known.
We solve the following two optimization problems, changing parameters $\gamma\in[0,1]$ and $\eta\in[-1,+1]$.
\begin{equation}
  \begin{aligned}
    \min_{X} \quad & L_{\mathrm{PUNU}}^{\gamma}(X,A)\\
    \st \quad & X \in K_{\max}(\alpha, \alpha\sqrt{r}).
  \end{aligned}
\end{equation}
\begin{equation}
  \begin{aligned}
    \min_{X} \quad & L_{\mathrm{PNU}}^{\eta}(X,A)\\
    \st \quad & X \in K_{\max}(\alpha, \alpha\sqrt{r}).
  \end{aligned}
\end{equation}
We denote our estimated matrix by $M^{*}$.

\begin{figure}[tbp]
  \begin{center}
    \includegraphics[width=0.7\linewidth]{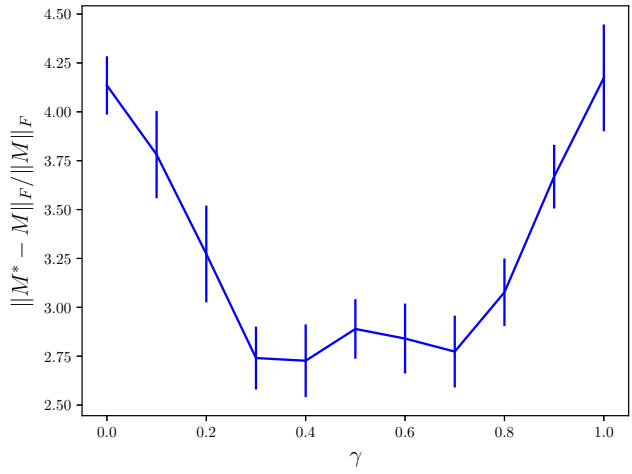}
    \caption{Plot of the relative Frobenius error
      $\|M^{*}-M\|_{\mathrm{F}} / \|M\|_{\mathrm{F}}$
      versus $\gamma$ of the PUNU risk.}
    \label{fig:exp1-1}
  \end{center}
\end{figure}

\begin{figure}[tbp]
  \begin{centering}
    \includegraphics[width=0.7\linewidth]{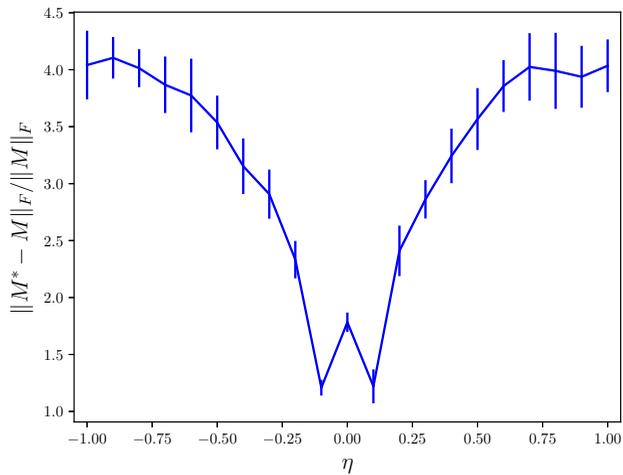}
    \caption{Plot of the relative Frobenius error
      $\|M^{*}-M\|_{\mathrm{F}} / \|M\|_{\mathrm{F}}$
      versus $\eta$ of the PNU risk.}
    \label{fig:exp1-2}
  \end{centering}
\end{figure}

We plot the average error, which is measured by relative Frobenius error $\|M^{*}-M\|_{\mathrm{F}} / \|M\|_{\mathrm{F}}$, and the standard deviation over 10 trials.
Fig.~\ref{fig:exp1-1} shows the result of the PUNU risk.
The left-most and right-most points correspond to the PU and NU risks, respectively.
In this experiment, $\gamma\in[0.3,0.7]$ works best, and we can see that the mixture of PU and NU risks improves the performance.
Fig.~\ref{fig:exp1-2} shows the result of the PNU risk.
The left-most ($\eta=-1$), middle ($\eta=0$) and right-most ($\eta=+1$) points correspond to the NU, PN and PU risks, respectively.
In this experiment, $\eta = \pm 0.1$ works the best.
Comparing the results of the two experiments, we can see that the PNU method is superior to the PUNU method if parameters are properly tuned, similarly to the result in the classification field \citep{Sakai17}.

\subsection{Synthetic data}
Second, we conduct a synthetic experiment to illustrate the behavior of the proposed risk $L_{\mathrm{TRI}}$.
We generate data in the same way as the previous experiment, with parameters $d_1=d_2=300$, $r=10$, $\alpha=1$, and $\rho=0.85$, using $f(x) = \frac{1}{(1+\exp(-7x))}$ for the QPF.
We again assume that these parameters are known.
We solve the optimization problem in Eq.~\eqref{opt:tri}, changing hyperparameters $\gamma_{\mathrm{PN}}$, $\gamma_{\mathrm{PU}}$, and $\gamma_{\mathrm{NU}}$.

\begin{figure}[tbp]
  \begin{center}
    \includegraphics[width=\linewidth]{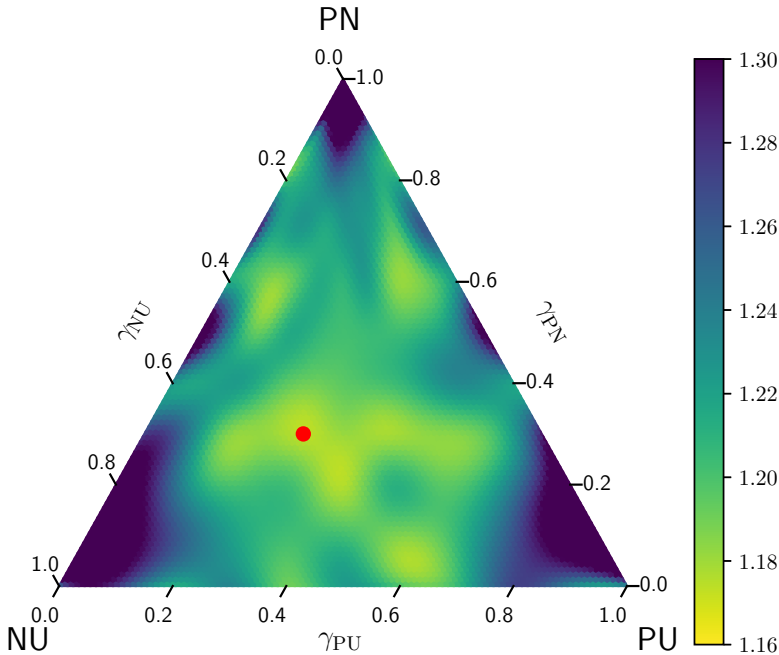}
    \caption{Ternary heatmap of the relative Frobenius error
      $\|M^{*}-M\|_{\mathrm{F}} / \|M\|_{\mathrm{F}}$ for synthetic data.}
    \label{fig:exp2}
  \end{center}
\end{figure}

We plot the average error over 10 trials in Fig.~\ref{fig:exp2}.
The best point, indicated by the red dot, is located around $(\gamma_{\mathrm{PN}}, \gamma_{\mathrm{PU}}, \gamma_{\mathrm{NU}}) \sim (0.3, 0.3, 0.4)$.
Since the PUNU and PNU methods can search only points on the edges of this plot, this result indicates that our method can be superior to those methods in the sense that it can search the inside of this triangular region.

\subsection{Real-world data}
Finally, we evaluate the performance of our method with real-world data.
We use the MovieLens (100k) dataset \citep{movie100k}.
This dataset contains 100,000 movie ratings from 943 users on 1682 movies.
Each rating has an integer value from 1 to 5.
Since we consider the binary matrix completion, we threshold them by the average value of all ratings.
The average is around 3.5, thus we transform ratings 1, 2, and 3 into $-1$, and ratings 4 and 5 into $+1$.
We keep 5,000 samples for validation and another 5,000 samples for testing, and then solve the optimization problem in Eq.~\eqref{opt:tri} using the remaining 90,000 samples, with the QPF $f(x) = 1 / (1+\exp(-x))$.
In this experiment, we have to estimate parameters other than $\gamma_{\mathrm{PN}}$, $\gamma_{\mathrm{PU}}$, and $\gamma_{\mathrm{NU}}$, that is, $\alpha$ and $r$.
Since it is computationally too expensive to tune all of them at once, we first estimate the value of $\alpha$ and $r$ using the PN risk, and then tune $\gamma_{\mathrm{PN}}$, $\gamma_{\mathrm{PU}}$, and $\gamma_{\mathrm{NU}}$.

Since only ratings which are already quantized are available, there is no way to measure the accuracy against the unknown underlying matrix.
Here we evaluate the estimated matrix by its accuracy on the prediction of the sign of the validation samples.
That is, we first quantize it comparing each entry with the average value, and then evaluate how accurately the signs of validation samples are predicted.

\begin{figure}[tbp]
  \begin{center}
    \includegraphics[width=\linewidth]{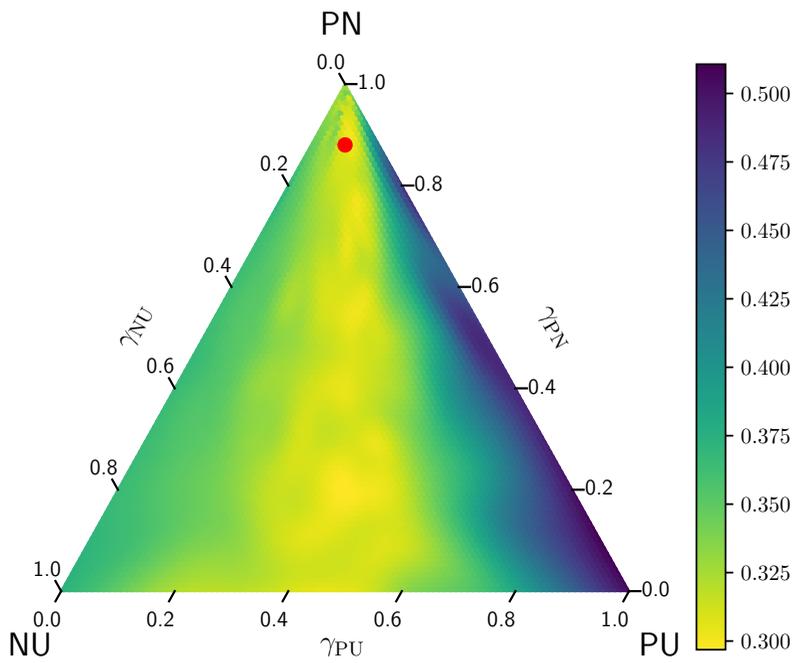}
    \caption{Ternary heatmap of the test misclassification error for the MovieLens dataset.}
    \label{fig:exp3-1}
  \end{center}
\end{figure}

Fig.~\ref{fig:exp3-1} shows the ternary heatmap of the error on test samples.
The best point, indicated by the red dot, again is located inside the triangular region.
Table~\ref{tb:exp3-2} shows the average error on validation samples of 10 trials of the PN and proposed methods with the best parameters.
Since we consider the max norm constraint, this PN method corresponds to the method of \citet{Cai13}.
Our method achieved 2\% lower error overall.
More precisely, our method achieves higher performances on negatively quantized samples, i.e., 1, 2, and 3, while the PN method works better on positively quantized samples, i.e., 4 and 5.
It is interesting that both methods perform poorly when the true rating is close to the average value.
From the viewpoint of the BMC problem setting, this is a natural phenomenon since the value of QPF becomes close to 0.5 and thus the quantization process becomes nearly at random.
Overall, this result shows that a proper mixture of the PN, PU, and NU risks can improve the performance also in the real-world problem, and supports the usefulness of our method.

\begin{table}[tbp]
  \centering{
    \caption[]{Misclassification error on validation samples. Averages and standard deviations of 10 trials.
      Boldface denotes the best according to the t-test at a significance level of 1\%.}\label{tb:exp3-2}
    \begin{tabular}{lccc} \hline
      Original rating & 1 & 2 & 3\\ \hline
      PN method & 0.303$\pm$0.034 & 0.298$\pm$0.026 & 0.491$\pm$0.015\\
      Proposed method & {\bf 0.219$\pm$0.019} & {\bf 0.245$\pm$0.021} & {\bf 0.434$\pm$0.023}\\
      \hline \hline
      Original rating & 4 & 5 & Overall\\ \hline
      PN method & {\bf 0.297$\pm$0.014} & 0.157$\pm$0.025 & 0.320$\pm$0.015\\
      Proposed method & 0.327$\pm$0.020 & 0.161$\pm$0.011 & {\bf 0.304$\pm$0.005}\\ \hline
    \end{tabular}
  }
\end{table}

\section{Conclusion}\label{ch:conclusion}
In this paper, we studied the binary matrix completion problem, proposed by \citet{Daven14}, where observations are quantized.
We first adapted the method of \citet{Hsieh15}, which is developed for the PU matrix completion problem, to the BMC setting.
Then we constructed the proposed method by combining it with the risk used in \citet{Daven14}.
Our method can handle unobserved entries, which the previous method did not utilize, in addition to the observed entries by tuning hyperparameters.
As far as we know, this is the first BMC method which can exploit all of the positive, negative, and unobserved entries.

The idea of combining risks is motivated by the semi-supervised classification method of \citet{Sakai17}.
However, we experimentally observed that the optimal mixture of risks for the BMC problem is different from that for the classification problem.
In experiments, we demonstrated that with both synthetic and real-world data, the optimal mixture of risks tends to consist all of the PN, PU, and NU risks, and thus our method is superior to previous methods.

Although our method worked better experimentally, we did not have theoretical guarantees on its performance.
Thus the theoretical analysis such as an upper bound on the recovery error would be an important direction for the future work.

Also, since we have parameters $\gamma_{\mathrm{PN}}$, $\gamma_{\mathrm{PU}}$, and $\gamma_{\mathrm{NU}}$ to be tuned, our method is computationally inefficient.
If we can develop either a heuristic or theoretical way to find optimal values of these parameters efficiently, it would make our method more practical.

\subsection*{Acknowledgement}
We would like to thank Issei Sato and Junya Honda for their support.
TS was supported by KAKENHI $15$J$09111$.
MS was supported by JST CREST JPMJCR1403.

\bibliographystyle{plainnat-reversed}
\bibliography{contents/myref}

\end{document}